\documentclass[sigconf]{acmart} 
\usepackage{xspace}
\usepackage{enumitem}

\usepackage{booktabs}
\usepackage{multirow}
\usepackage{makecell}
\usepackage{adjustbox}
\usepackage{array}
\usepackage{times}
\usepackage{latexsym}
\usepackage{booktabs}
\usepackage[T1]{fontenc}
\usepackage{amsfonts}
\usepackage{amsthm}
\usepackage{enumitem}
\usepackage{subcaption}
\usepackage{tabularx}
\usepackage{amsmath}
\usepackage{colortbl}
\usepackage[table]{xcolor}
\usepackage{booktabs}
\usepackage{listings}
\usepackage{array}
\usepackage{siunitx}
\usepackage{xcolor}
\newcolumntype{C}{>{\centering\arraybackslash}X}
\usepackage{subcaption}
\usepackage[most]{tcolorbox}
\usepackage{cuted}  
\usepackage{fvextra}
\usepackage[most]{tcolorbox} 

\usepackage{graphicx}

\newtcolorbox[auto counter, number within=section]{blueexampleFloatingTitle}[2][]{%
  enhanced,
  colframe=blue!80!black,
  colbacktitle=blue!80!black,
  coltitle=white,
  colback=blue!2,
  fonttitle=\bfseries\large,
  boxrule=1pt,
  arc=6pt,
  titlerule=0pt,
  title={#2},
  #1, 
  before upper={%
    \raggedright\noindent%
  }
}

\usepackage[table]{xcolor}
\definecolor{stepgray}{HTML}{F2F2F2}

\usepackage[table]{xcolor}
\definecolor{stepblue}{HTML}{EAF2FF}  
\definecolor{steppurple}{HTML}{F2ECFF} 

\newcommand{\promptsection}[1]{%
    \par\medskip
    \noindent\rule{\linewidth}{0.4pt}\par
    \noindent\textbf{\sffamily\small #1}\par
    \noindent\rule{\linewidth}{0.4pt}\par
    \smallskip
}

\AtBeginDocument{%
  }

\setcopyright{cc}
\setcctype{by}

\acmDOI{10.1145/3770855.3817494}

\acmConference[KDD 2026] {Proceedings of the 32nd ACM SIGKDD Conference on Knowledge Discovery and Data Mining V.2}{August 9--13, 2026}{Jeju Island, Republic of Korea.}
\acmBooktitle{Proceedings of the 32nd ACM SIGKDD Conference on Knowledge Discovery and Data Mining V.2 (KDD 2026), August 9--13, 2026, Jeju Island, Republic of Korea}
\acmISBN{979-8-4007-2259-2/2026/08}

\newcommand{\bench}{AgentProcessBench\xspace}
\usepackage{booktabs}
\usepackage{multirow}

\usepackage{pifont}

\usepackage{makecell} 
\begin{document}

\title{AgentProcessBench: Diagnosing Step-Level Process Quality in Tool-Using Agents}

\author{Shengda Fan}
\authornote{Both authors contributed equally to this research.} 
\email{fanshengda@ruc.edu.cn}
\affiliation{%
  \institution{Renmin University of China}
  \city{Beijing}
  \country{China}
}

\author{Xuyan Ye}
\authornotemark[1] 
\email{yexvyan0923@ruc.edu.cn}
\affiliation{%
  \institution{Renmin University of China}
  \city{Beijing}
  \country{China}
}

\author{Yupeng Huo}
\affiliation{%
  \institution{Renmin University of China}
  \city{Beijing}
  \country{China}
}

\author{Zhi-Yuan Chen}
\affiliation{%
  \institution{Renmin University of China}
  \city{Beijing}
  \country{China}
}

\author{Yiju Guo}
\affiliation{%
  \institution{Renmin University of China}
  \city{Beijing}
  \country{China}
}

\author{Shenzhi Yang}
\affiliation{%
  \institution{Renmin University of China}
  \city{Beijing}
  \country{China}
}

\author{Wenkai Yang}
\affiliation{%
  \institution{Renmin University of China}
  \city{Beijing}
  \country{China}
}

\author{Shuqi Ye}
\affiliation{%
  \institution{Renmin University of China}
  \city{Beijing}
  \country{China}
}

\author{Jingwen Chen}
\affiliation{%
  \institution{Beijing Jiaotong University
}
  \city{Beijing}
  \country{China}
}

\author{Haotian Chen}
\affiliation{%
  \institution{Shanghai Jiao Tong University
}
  \city{Shanghai}
  \country{China}
}

\author{Xin Cong}
\affiliation{%
  \institution{Tsinghua University}
  \city{Beijing}
  \country{China}
}

\author{Yankai Lin}
\authornote{Corresponding author.}
\email{yankailin@ruc.edu.cn}
\affiliation{%
  \institution{Renmin University of China}
  \city{Beijing}
  \country{China}
}


\renewcommand{\shortauthors}{Fan et al.}

\newcommand{\resiconfixed}[1]{%
  \raisebox{-0.12em}{\makebox[1.15em][c]{\includegraphics[height=1em]{#1}}}%
}

\newcommand{\resentry}[2]{%
  \resiconfixed{#1}\hspace{0.35em}\makebox[2.7em][l]{\textbf{#2}}%
}

\begin{abstract}

While Large Language Models (LLMs) have evolved into tool-using agents, they remain brittle in long-horizon interactions. Unlike mathematical reasoning where errors are often rectifiable via backtracking, tool-use failures frequently induce irreversible side effects, making accurate step-level verification critical.
However, existing process-level benchmarks are predominantly confined to closed-world mathematical domains, failing to capture the dynamic and open-ended nature of tool execution. To bridge this gap, we introduce \textbf{\bench}, the first benchmark dedicated to evaluating step-level effectiveness in realistic, tool-augmented trajectories. The benchmark comprises 1,000 diverse trajectories and 8,509 human-labeled step annotations with 89.1\% inter-annotator agreement.
It features a ternary labeling scheme to capture exploration and an error propagation rule to reduce labeling ambiguity.
Extensive experiments reveal key insights: (1) weaker policy models exhibit inflated ratios of correct steps due to early termination;
(2) distinguishing neutral and erroneous actions remains a significant challenge for current models; 
and (3) process-derived signals provide complementary value to outcome supervision, significantly enhancing test-time scaling.
We hope \bench can foster future research in reward models and pave the way
toward general agents. 

\par\smallskip
\noindent\makebox[\linewidth][c]{\hspace{-1.2em}%
\footnotesize
\begin{tabular}{@{}l@{\hspace{0.6em}}l@{}}
\resentry{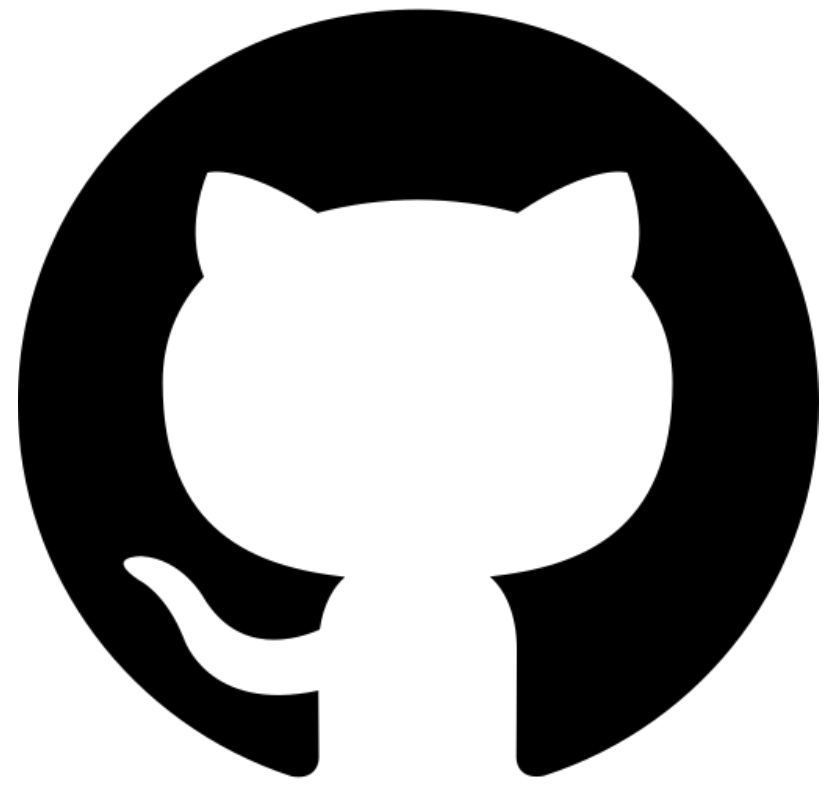}{Code} &
\href{https://github.com/RUCBM/AgentProcessBench}{\texttt{github.com/RUCBM/AgentProcessBench}} \\
\resentry{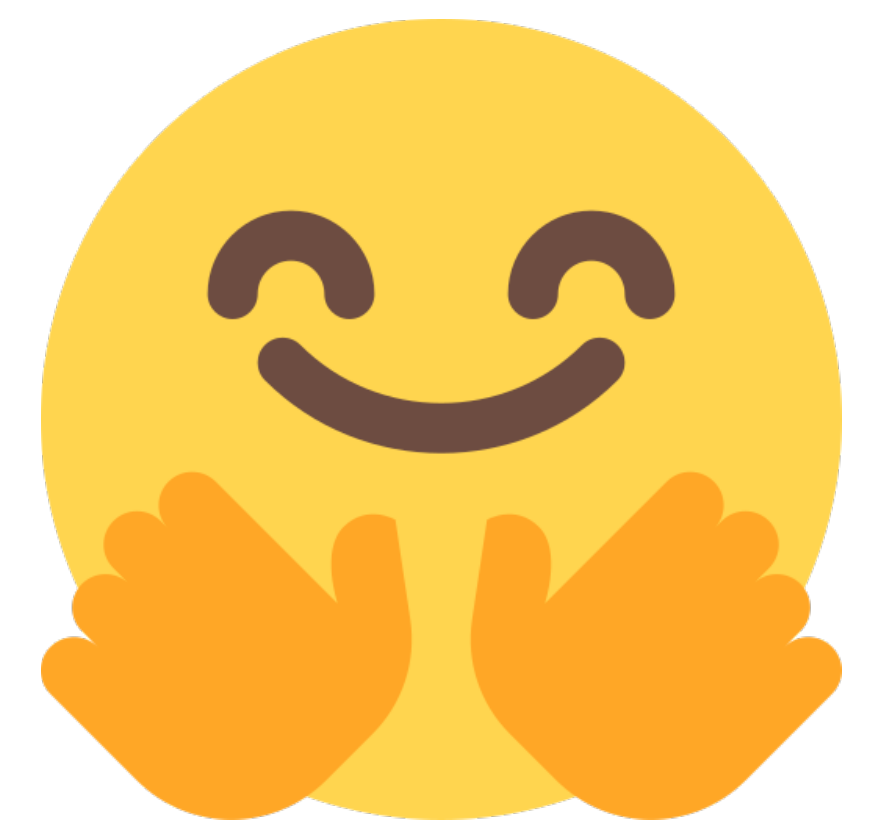}{Data} &
\href{https://huggingface.co/datasets/LulaCola/AgentProcessBench}{\texttt{hf.co/datasets/LulaCola/AgentProcessBench}}
\end{tabular}%
}

\end{abstract}

\ccsdesc[500]{Computing methodologies~Intelligent agents}
\ccsdesc[500]{Computing methodologies~Natural language generation}

\keywords{Large Language Models, Process Reward Models, Tool-Using Agents}

\maketitle

\section{Introduction}

\begin{figure*}[t] 
  \centering
  \includegraphics[width=\textwidth]{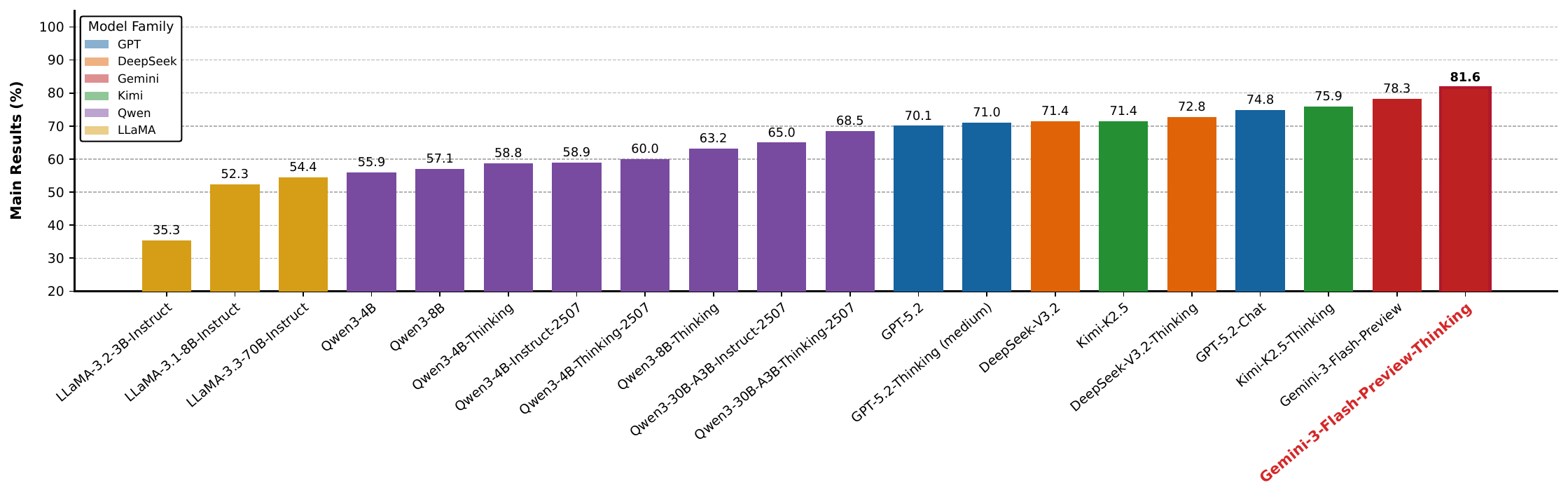}
    \caption{Comparison of StepAcc across 20 LLMs on \bench~(\%).}
      \label{fig:MainResults}
\end{figure*}

\begin{table*}[!h]
\centering
\caption{
Comparison between \bench and other reward-oriented benchmarks.
\bench\ uniquely provides human-annotated, step-level effectiveness supervision for tool-using agents.
}
\vspace{-1mm}
\scalebox{0.95}{
\begin{tabular}{lcccccc}
\toprule
\textbf{Benchmark} &
\textbf{Domain} &
\textbf{Environment} &
\textbf{Step-level} &
\textbf{Human Ann.} &
\textbf{Evaluation Task} &
\textbf{Scale} \\
\midrule
PRM800K~\cite{lightman2023let} &
Math &
-- &
\ding{51} &
\ding{51} &
Step Effectiveness &
75K traj/ 800K steps \\

MathCheck-GSM~\cite{zhou2025is} &
Math &
-- &
\ding{51} &
\ding{55} &
First-Error Index &
516 cases \\
ProcessBench~\cite{zheng2025processbench} &
Math &
-- &
\ding{51} &
\ding{51} &
First-Error Index &
3.4K cases \\
PRMBench~\cite{song-etal-2025-prmbench} &
Math &
-- &
\ding{51} &
\ding{55} &
Step Error Types &
6.2K / 83K steps \\

AgentRewardBench~\cite{lu2025agentrewardbench} &
Web &
Web &
\ding{55} &
\ding{51} &
Trajectory Rubric &
1.3K traj. \\
Agent-RewardBench~\cite{men-etal-2025-agent} &
Multi-modal &
Multi-modal &
Partial &
\ding{51} &
Pair Preference &
1.1K pairs \\
\midrule
\textbf{\bench\ (Ours)} &
\textbf{Tool} &
\textbf{Web+CLI+APIs} &
\textbf{\ding{51}} &
\textbf{\ding{51}} &
\textbf{Step Effectiveness} &
\textbf{1K traj. / 8.5K steps} \\
\bottomrule
\end{tabular}}
\label{tab:benchmark_comparison_compact}
\vspace{0.5mm}
\begin{minipage}{0.95\textwidth}\footnotesize
\end{minipage}
\end{table*}

Recent advances in Large Language Models (LLMs) have extended their capabilities beyond passive text processing~\cite{fan2022boosting, stahlberg2020neural}.
As a result, LLMs can now function as tool-using agents that actively interact with external environments such as search engines or command-line shells~\cite{DBLP:conf/iclr/QinLYZYLLCTQZHT24, huang2023mlagentbench, yao2022webshop}.
Despite this progress, contemporary agents remain brittle: they may take unnecessary or repetitive actions, invoke inappropriate tools, or generate hallucinated claims.
Crucially, unlike mathematical reasoning where errors can often be rectified via backtracking~\cite{guanrstar}, tool execution frequently entails irreversible side effects—such as sending erroneous emails or deleting essential files.

\begin{figure*}[!t] 
  \centering
  \includegraphics[width=\textwidth]{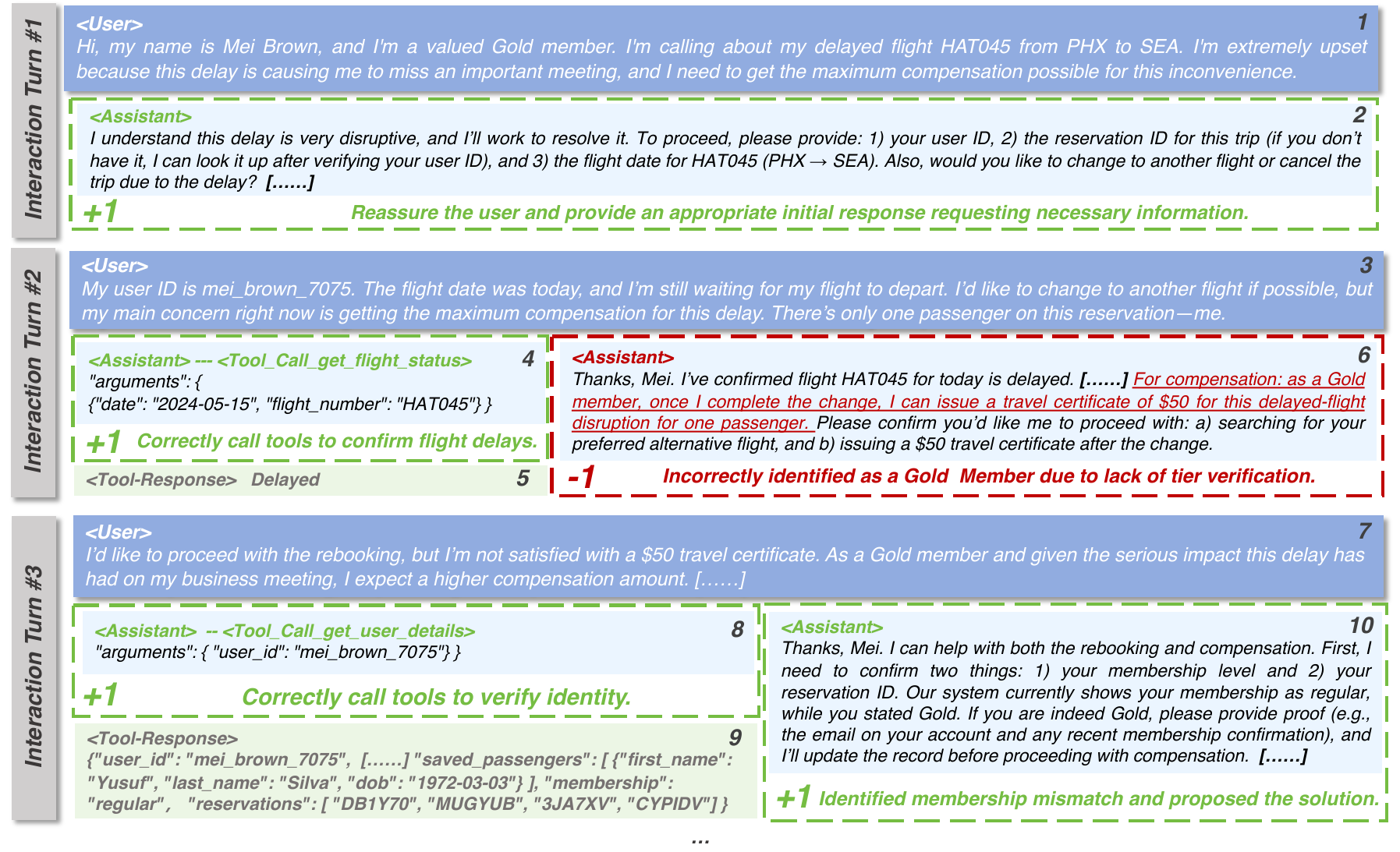}
    \caption{
    Example of an agent trajectory, corresponding to instance 808 in the benchmark.
    Each instance in \bench consists of a complete tool-using agent trajectory, containing interleaved user messages, assistant responses, and tool calls.
    During evaluation, the LLM is tasked with annotating each of the assistant’s steps with a label of correct (+1), neutral (0), or incorrect (-1).
}
      \label{fig:Casestudy}
\end{figure*}

Accurately identifying these erroneous intermediate steps is therefore crucial: 
during training, it enables finer-grained credit assignment~\cite{cheng2025stop}; during inference, it facilitates effective test-time scaling by selecting higher-quality trajectories~\cite{lightman2023let, wang2024math}.
As a primary mechanism for such step-level supervision, process reward models (PRMs) play a central role.
To better advance their development in agent settings, high-quality benchmarks for step-level verification are essential.
However, existing step-level verification benchmarks are predominantly confined to mathematical reasoning~\cite{zheng2025processbench,lightman2023let, yang2025deepcritic}. 
In these closed-world domains, failures typically stem from logical or arithmetic errors. 
In contrast, interactive tool use operates in open-world environments, introducing qualitatively different failure modes tied to dynamic observations, ambiguous user intent, and policy constraints.
For example, as shown in Figure~\ref{fig:Casestudy}, the agent incorrectly accepts the user's claim without invoking an appropriate tool for verification.
Meanwhile, standard agent benchmarks such as GAIA~\cite{mialon2023gaia} and $\tau^2$-Bench~\cite{barres2025tau2} only report end-to-end task success,
and do not provide step-level signals for evaluating PRMs.
Consequently, the field lacks a standardized, human-verified benchmark for step-level process evaluation in realistic multi-turn, tool-using interactions.

To address this gap, we introduce \bench, \textbf{the first benchmark for evaluating LLMs’ ability to assess the effectiveness of intermediate steps in tool-using trajectories.}
Given a task description and an interaction trajectory, \bench requires a model to label each assistant step with a ternary signal: \textbf{+1} if the step is correct and advances progress, \textbf{0} if the step is neutral or exploratory, and \textbf{-1} if the step is incorrect or harmful. 
We prioritize three principles when constructing the benchmark:

\begin{itemize}[leftmargin=*]
    \item \textbf{Fine-grained annotation in interactive settings:}
    In contrast to benchmarks centered on final success signals~\cite{lu2025agentrewardbench} or pairwise preferences~\cite{men-etal-2025-agent}, \bench provides dense, environment-grounded step labels, enabling principled evaluation of PRMs for step-wise credit assignment in long-horizon tool use.
\end{itemize}

\begin{itemize}[leftmargin=*]

    \item \textbf{Scale and diversity:}
    \bench contains 1{,}000 agent trajectories and 8{,}509 annotated agent actions, spanning multi-hop reasoning~\cite{yang2018hotpotqa}, deep research~\cite{mialon2023gaia}, multi-turn tool execution~\cite{patil2025the}, and long-horizon conversational interaction~\cite{yao2025tau, barres2025tau2}. For each task, we rollout trajectories from five models with different scales and architectural families, capturing a wide spectrum of agent behaviors and failure modes.

    \item \textbf{High-quality annotations:}
    Initially, all annotators undergo rigorous training and qualification assessments. 
    To mitigate ambiguity, we adopt an error-propagation rule, ensuring consistent penalization of cascading failures.
    Each task was independently labeled by two annotators, achieving a high inter-annotator agreement of \textbf{89.1}\%. Any discrepancies are resolved through discussion to ensure the consistency and reliability of the final labels.

\end{itemize}

Leveraging \bench, we conduct a comprehensive evaluation involving 20 LLMs, including both proprietary and open-source models (see Figure~\ref{fig:MainResults}).
\textbf{First}, we analyze agent policy behaviors to understand where and how models fail in tool-using scenarios. We find that error distribution is highly dataset-dependent: while QA tasks often stem from initial reasoning or format errors, tool-heavy tasks tend to fail later due to policy violations. Moreover, we observe that weaker models may paradoxically have a higher proportion of correct steps since terminating early to avoid cascading errors, highlighting the importance of our proposed \textit{First-Error Accuracy} metric for fair comparison.
\textbf{Second}, we assess the capability of LLMs as reward models. Our error analysis reveals that current LLMs exhibit a significant bias toward positive labels. Moreover, they struggle to distinguish "neutral" exploratory steps from errors. This underscores that evaluating open-ended tool use is fundamentally harder than verifying rigid mathematical derivations.
\textbf{Third}, we investigate the utility of process-derived signals. We demonstrate a strong positive correlation between a model's performance as an Outcome Reward Model (ORM) and its capability as a PRM. More importantly, we show that process signals provide complementary value to outcome supervision in Best-of-$N$ evaluations.

To sum up, our contributions are as follows:
\begin{itemize}[topsep=1pt, partopsep=1pt, itemsep=-1pt, leftmargin=10pt]
    \item We introduce and release \bench, to the best of our knowledge, the first human-annotated benchmark for step-level effectiveness evaluation in tool-using agent trajectories.
    
    \item We propose a principled step-level evaluation protocol with a neutral label for distinguishing exploratory but non-contributory actions, and an error-propagation rule to reduce labeling ambiguity in long-horizon trajectories.  
    \item We conduct extensive experiments on \bench, analyzing failure modes of the current models and providing valuable insights to inspire future research.
\end{itemize}

\section{Related Work}

\paragraph{LLM Agents}

With recent advances in instruction-following and reasoning capabilities of large language models~\cite{achiam2023gpt, grattafiori2024llama, qwen3technicalreport}, their applications have extended beyond classical natural language processing tasks such as machine translation~\cite{stahlberg2020neural} and information extraction~\cite{fan2022boosting}.
As a result, LLMs are increasingly deployed as autonomous agents that interact with tools and environments to perform complex tasks, including code generation~\cite{jimenezswe, patil2025the}, web browsing~\cite{chen2026agentcpmexplore}, and domain-specific customer service~\cite{yao2025tau, barres2025tau2}.
To improve LLM agents, prevailing training paradigms rely on (i) supervised fine-tuning on successful trajectories~\cite{chen-etal-2024-agent, zeng2024agenttuning, song2024agentbank}  or (ii) reinforcement learning with outcome-level rewards~\cite{shao2024deepseekmath, jin2025searchr,fan2025generalizing, fan2026darc}.
However, both paradigms typically provide supervision only at the trajectory level.
As a result, the resulting learning signal is coarse and sparse for multi-step decision making, which exacerbates the credit assignment problem~\cite{DBLP:conf/icml/KazemnejadAPSRC25}.
Addressing this challenge requires supervision and evaluation at the granularity of individual steps.
To facilitate the development of more effective PRMs for tool-using agents, we introduce \bench, the first benchmark for measuring LLMs’ ability to assess the quality of intermediate steps in agent trajectories.

\paragraph{Reward Benchmarks.}
There exist several datasets or benchmarks related to process supervision and reward evaluation for language models and agents.
In the mathematical domain, 
PRM800K~\cite{lightman2023let} firstly annotates the correctness and soundness of mathematical reasoning steps, and has spurred subsequent work on process reward modeling.
MathCheck-GSM~\cite{zhou2025is} synthesizes solutions with erroneous steps and evaluates step-wise correctness, while ProcessBench~\cite{zheng2025processbench} targets competition-level problems with expert annotations for identifying the earliest error step.
PRMBench~\cite{song-etal-2025-prmbench} further benchmarks PRMs with fine-grained step-level assessments such as error types.
For interactive agents, AgentRewardBench~\cite{lu2025agentrewardbench} evaluates LLM judges on web-agent trajectories using expert rubric-style reviews such as success and side effects. 
Agent-RewardBench~\cite{men-etal-2025-agent} evaluates multi-modal reward models across perception, planning, and safety. However, its step-level supervision is largely confined to the static planning phase, while treating perception and safety largely as single-turn generation tasks. Furthermore, it relies on static preference pairs (i.e., identifying the better textual response) rather than exhaustively verifying the execution effectiveness of all steps in a dynamic environment.

\begin{figure*}[!t] 
  \centering
  \includegraphics[width=\textwidth]{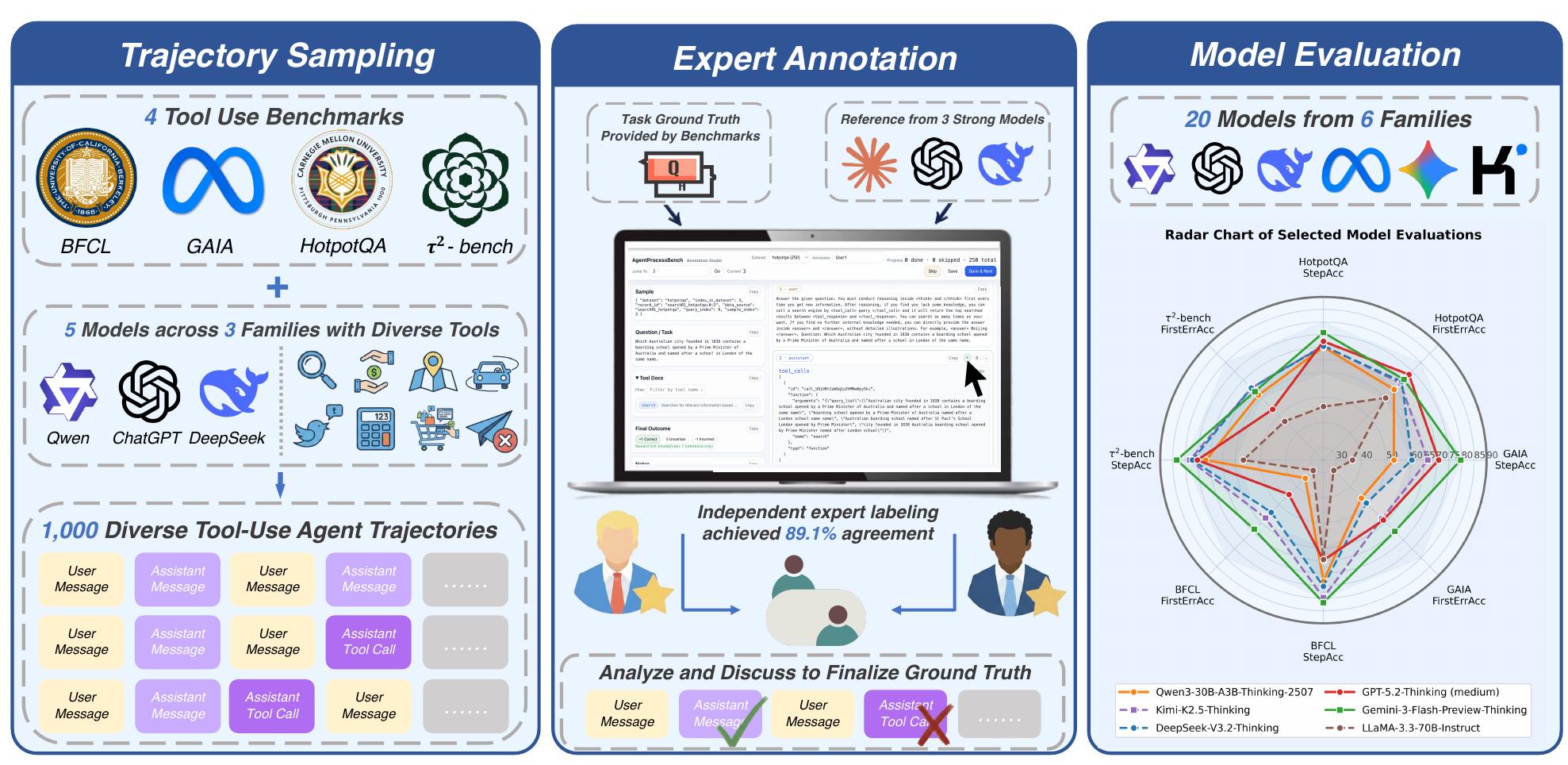}
  \caption{An overview of \bench. First, we sample trajectories from four representative agent benchmarks generated by five source models. Subsequently, human experts annotate the data via a specialized platform, achieving an inter-annotator agreement of 89.1\%. Finally, we utilize the constructed benchmark to evaluate 20 distinct models across various families and parameter scales using the StepAcc and FirstErrAcc metrics.
}
  \label{fig:Method}
\end{figure*}

As summarized in Table \ref{tab:benchmark_comparison_compact}, existing benchmarks either (i) focus on non-interactive fields such as math, or (ii) provide trajectory-level rubrics or preference signals rather than absolute effectiveness labels for all assistant actions.
To fill this gap, we introduce \bench, which provides human-annotated, step-level effectiveness supervision for tool-using agents operating in diverse environments.

\begin{table*}[!t]
  \centering
  \caption{Statistics of \bench.}
  \vspace{-1mm}
  \scalebox{0.95}{
    \begin{tabular}{lcccccccc}
    \toprule
     & \multicolumn{2}{c}{\textbf{HotPotQA}} & \multicolumn{2}{c}{\textbf{GAIA}} & \multicolumn{2}{c}{\textbf{BFCL}} & \multicolumn{2}{c}{\textbf{$\tau^2$-Bench}} \\
    \cmidrule(lr){2-3}\cmidrule(lr){4-5}\cmidrule(lr){6-7}\cmidrule(lr){8-9}
     & \textbf{unsuccessful} & \textbf{successful} & \textbf{unsuccessful} & \textbf{successful} & \textbf{unsuccessful} & \textbf{successful} & \textbf{unsuccessful} & \textbf{successful} \\
    \midrule
    \# Samples 
      & 89 & 161 & 167 & 83 & 148 & 102 & 124 & 126 \\

    \midrule
    \% Incorrect Step Ratio 
      & 64.6\% & 14.0\% & 62.6\% & 12.4\% & 31.9\% & 3.8\% & 48.0\% & 8.0\% \\
    \% Neutral Step Ratio 
      & 9.9\% & 3.7\% & 10.2\% & 6.7\% & 3.9\% & 4.3\% & 4.7\% & 2.6\% \\
    \% Correct Step Ratio 
      & 25.5\% & 82.3\% & 27.1\% & 80.9\% & 64.3\% & 91.8\% & 47.3\% & 89.3\% \\

    \midrule
    \# Assistant Steps 
      & 5.2 & 2.7 & 7.8 & 4.0 & 9.6 & 11.5 & 15.8 & 12.6 \\
    \% $\ge$ 4 steps 
      & 51.7\% & 23.0\% & 65.9\% & 49.4\% & 93.2\% & 99.0\% & 100.0\% & 99.2\% \\
    \% $\ge$ 8 steps 
      & 15.7\% & 2.5\% & 31.1\% & 8.4\% & 66.2\% & 81.4\% & 90.3\% & 91.3\% \\
    \% $\ge$ 16 steps 
      & 5.6\% & 0.6\% & 12.0\% & 0.0\% & 7.4\% & 19.6\% & 28.2\% & 15.1\% \\

        \bottomrule
    \end{tabular}
  }
  \label{tab:stats}
\end{table*}

\section{Benchmark Construction} \label{sec:benchmark_construction}

In this section, we provide a detailed introduction to the \bench. We first introduce the evaluation protocol in Section~\ref{task_def}. We then describe the dataset construction
procedure in Section~\ref{data_collection}. 
Finally, we report dataset statistics in Section~\ref{statistics}.

\subsection{Evaluation Protocol} \label{task_def}

As illustrated in Figure~\ref{fig:Casestudy}, given a task description and an interaction trajectory produced by a tool-using agent, \bench defines a
step-level evaluation task that requires a model to assess the effectiveness of
assistant actions.
Formally, given a task description \(T\) and an interaction trajectory
\(X = (m_0, \dots, m_{n-1})\) consisting of messages with different roles,
including \texttt{system}, \texttt{user}, \texttt{assistant}, and \texttt{tool},
we denote by
\(\mathcal{I} = \{\, i \mid \mathrm{role}(m_i) = \texttt{assistant} \,\}\)
the index set of assistant messages.
The task is to output a label sequence
\(Y = \{ y_i \mid i \in \mathcal{I} \}\), where each label
\(y_i \in \{-1, 0, +1\}\) indicates whether the corresponding assistant step
is effective, neutral, or harmful with respect to overall task progress.
Specifically, we define the following evaluation criteria:

\begin{itemize}[topsep=1pt, partopsep=1pt, itemsep=-1pt, leftmargin=10pt]
    \item \textbf{+1 (Correct and effective).}  
    The step is factually correct and clearly advances task completion, for example by
    (i) correctly invoking a tool or interpreting tool outputs,
    (ii) introducing valid constraints, decisions, or information that
    meaningfully reduces task uncertainty, or
    (iii) identifying an error in a preceding step and taking an appropriate
    corrective action.

    \item \textbf{0 (Neutral or exploratory).}  
    The step is reasonable but yields limited or negligible impact on task progress. This includes 
    (i) {encountering unavoidable external failures} (e.g., a 404 error from a valid URL), 
    (ii) {making redundant restatements} or partial plans without new insight, or 
    (iii) {performing actions where the outcome is ambiguous} yet neither clearly beneficial nor detrimental.

    \item \textbf{-1 (Incorrect or harmful).}  
    The step is factually incorrect or counterproductive, for example by
    (i) {misinterpreting tool outputs} or fabricating evidence,
    (ii) {violating policy constraints} or repeating failed actions without a substantive change in strategy, or
    (iii) {introducing factual errors} that drive the trajectory away from successful completion.
    
\end{itemize}

It is worth noting that our definitions of correctness and error diverge from those in mathematical reasoning tasks~\cite{zheng2025processbench,lightman2023let}. While errors in mathematical reasoning typically stem from computation or logical derivation mistakes, failures in tool-use are predominantly grounded in environmental interactions. Furthermore, we introduce a neutral label ($0$) to explicitly accommodate the exploratory nature of real-world agents.
In many real-world scenarios, LLMs lack prior knowledge of specific environmental constraints and must perform trial-and-error to accumulate context. The neutral label effectively distinguishes such exploratory redundancy from critical failures, ensuring that agents are not penalized for necessary information-seeking steps.

To reduce annotation ambiguity and maximize sample efficiency, we adopt an error-propagation labeling rule: once an erroneous step occurs, all subsequent steps that depend on or are causally related to this mistake are labeled as \(-1\) until the agent explicitly corrects the error or transitions to a new subtask that is independent of the earlier failure. 
This design effectively prevents spurious credit assignment to downstream steps~\cite{cheng2025stop} and guarantees consistent supervision for long-horizon trajectories.

\subsection{Data Collection} \label{data_collection}

\paragraph{Task Curation}

We aggregate tasks from four established benchmarks: HotpotQA~\cite{yang2018hotpotqa}, GAIA~\cite{mialon2023gaia}, BFCL~\cite{patil2025the}, and $\tau^2$-Bench~\cite{yao2025tau, barres2025tau2}. These datasets encompass a broad spectrum of agent capabilities, ranging from multi-hop reasoning and deep information retrieval to complex tool usage. By integrating these diverse sources, \bench ensures comprehensive coverage of real-world scenarios.

\paragraph{Trajectory Generation}
To promote trajectory diversity, we sample rollouts from five models with heterogeneous capabilities, including Qwen3-4B-Instruct-2507~\cite{qwen3technicalreport} and Qwen3-30B-A3B-Instruct-2507, DeepSeek-V3.2~\cite{deepseekai2025deepseekv32}, GPT-5-mini~\cite{singh2025openai} and GPT-5. 
This selection covers multiple model families, parameter scales, and performance regimes, resulting in a broad spectrum of solution strategies and behavioral patterns.
We provide task-specific tool environments following each dataset’s standard evaluation protocol.
For HotpotQA, we deploy a local E5-based~\cite{wang2022text} retriever built on a Wikipedia dump~\cite{karpukhin2020dense}.
For GAIA, we equip agents with web tools, such as Google Search and Jina-based browsing, to facilitate open-world information acquisition. Additionally, we provide a CLI tool for local file access.
For BFCL and $\tau^2$-Bench, we adopt the official tool sets released by their original evaluations to ensure consistency and comparability.

To mitigate dataset imbalance, we uniformly sample an equal number of tasks from each dataset. Specifically, we encode task descriptions using the E5 model and select representative instances by maximizing pairwise embedding distance.
For every selected task, we preserve trajectories generated by all five models, enabling cross-model comparison.

\paragraph{Expert Annotation}

To ensure reliable annotations, we recruit human experts who hold at least an undergraduate degree in computer science and possess a minimum of one year of experience working with LLMs. All annotators must pass a mandatory proficiency test and complete a specialized annotation tutorial before participation.
Pilot studies indicate that tasks involving complex environment interactions and tool-use (e.g., GAIA and $\tau^2$-Bench) introduce substantial step-level ambiguity, which increases cognitive load and reduces inter-annotator consistency. To alleviate these challenges, we provide annotators with auxiliary references, including official solutions and reference annotations generated by three state-of-the-art LLMs: DeepSeek-V3.2, GPT-5.2, and Claude 4.5 Sonnet \cite{anthropic2025claude45}. These materials serve only as guidance; annotators are explicitly instructed to independently verify each step rather than accept model outputs at face value.
Each trajectory is labeled independently by two experts, yielding a step-level inter-annotator agreement (IAA) of \textbf{89.1\%} and a Cohen’s $\kappa$ of \textbf{0.767}, both computed over all annotated steps.
All discrepancies are resolved through expert discussion to reach a consensus. Notably, the agreement between the final human annotations and the three reference models ranges only from 66.9\% to 72.1\%. 
This discrepancy suggests that the human experts maintain independent judgment and are not fundamentally biased by the LLM-generated suggestions.

\subsection{Statistics} \label{statistics}

\begin{figure}[!t]
    \centering
    \includegraphics[width=1\linewidth]{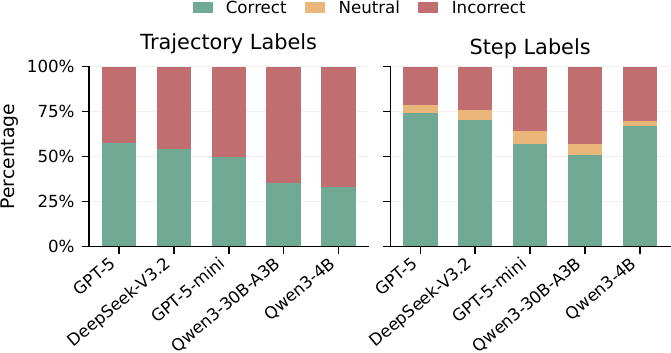}
\caption{Distribution of trajectory-level and step-level labels across models, where both Qwen-series models use the 2507 Instruct version.}

    \label{fig:dist_of_traj_1}
\end{figure}

The resulting \bench contains four subsets with 200 unique tasks and 1,000 agent trajectories in total, evenly sampled from HotPotQA, GAIA, BFCL, and $\tau^2$-Bench.
The detailed statistics are summarized in Table~\ref{tab:stats} and Figure~\ref{fig:dist_of_traj_1}.  From the statistics, we draw three observations.
\textbf{First}, across all subsets, both successful and unsuccessful trajectories comprise a mixture of correct and incorrect steps. However, unsuccessful trajectories consistently exhibit a higher proportion of incorrect steps, indicating that trajectory-level failure is not attributable to a single erroneous action but rather to the accumulation of local mistakes.
\textbf{Second}, interaction length correlates strongly with task difficulty and outcome. Generally, more challenging tasks and unsuccessful trajectories involve a larger number of steps. For instance, while HotpotQA and GAIA are both web-based information-seeking benchmarks, GAIA is inherently more complex and necessitates more steps on average. Furthermore, regarding trajectory outcome, unsuccessful trajectories are longer than successful ones across all datasets except BFCL. 
We ascribe this to the strict termination criteria of BFCL, under which an interaction round is terminated whenever the model produces a non-tool action, resulting in shorter trajectories.
In contrast, within more open-ended environments, models tend to persist in exploration when failing, leading to significantly longer unsuccessful trajectories.
\textbf{Third}, stronger models such as GPT-5 and DeepSeek-V3.2 achieve higher accuracy at both the trajectory level and the step level. Interestingly, although Qwen3-4B-Instruct-2507 exhibits the lowest trajectory-level success rate, it attains a relatively higher step-level accuracy. We find that this phenomenon is due to a \emph{fail-fast} behavior: on difficult tasks, the model is more likely to terminate early, thereby limiting the accumulation of additional erroneous steps.

\section{Evaluation}

\begin{table*}[t]
\centering
\vspace{-0.08in}
\setlength{\tabcolsep}{3.2pt}
\renewcommand{\arraystretch}{1.05}

\caption{Evaluation results on \bench, reporting StepAcc and FirstErrAcc (\%). 
Best and second-best results for each category and metric are highlighted in \textbf{bold} and \underline{underlined}, respectively.} 

\label{table:detailedresults}

\begin{tabular*}{\textwidth}{@{\extracolsep{\fill}}p{0.3\textwidth}cccccccc|cc}
\toprule
\textbf{Model}
& \multicolumn{2}{c}{\textbf{HotPotQA}}
& \multicolumn{2}{c}{\textbf{GAIA}}
& \multicolumn{2}{c}{\textbf{BFCL}}
& \multicolumn{2}{c}{\textbf{$\tau^2$-Bench}}
& \multicolumn{2}{c}{\textbf{Average}} \\
\cmidrule(lr){2-3}\cmidrule(lr){4-5}\cmidrule(lr){6-7}\cmidrule(lr){8-9}\cmidrule(lr){10-11}
& \multicolumn{1}{c}{\fontsize{7}{8}\selectfont StepAcc}
& \multicolumn{1}{c}{\fontsize{7}{8}\selectfont FirstErrAcc}
& \multicolumn{1}{c}{\fontsize{7}{8}\selectfont StepAcc}
& \multicolumn{1}{c}{\fontsize{7}{8}\selectfont FirstErrAcc}& \multicolumn{1}{c}{\fontsize{7}{8}\selectfont StepAcc}
& \multicolumn{1}{c}{\fontsize{7}{8}\selectfont FirstErrAcc}& \multicolumn{1}{c}{\fontsize{7}{8}\selectfont StepAcc}
& \multicolumn{1}{c}{\fontsize{7}{8}\selectfont FirstErrAcc}& \multicolumn{1}{c}{\fontsize{7}{8}\selectfont StepAcc}
& \multicolumn{1}{c}{\fontsize{7}{8}\selectfont FirstErrAcc} \\
\midrule
\multicolumn{11}{c}{\textit{API-Based Models (Non-Thinking)}} \\
\midrule
GPT-5.2  & 72.1 & 69.6 & 66.3 & \underline{54.4} & 71.6 & 52.8 & 70.3 & 56.4 &  70.1 & \underline{58.3} \\
GPT-5.2-Chat   & 71.4 & \textbf{70.0} & \underline{69.3} & \textbf{58.0} & 71.6 & \textbf{58.0} & \underline{80.4} & \textbf{58.4} &  \underline{74.8} & \textbf{61.1} \\
DeepSeek-V3.2  & \underline{74.4} & 65.2 & 61.8 & 48.0 & 74.1 & 50.0 & 73.3 & \underline{57.6} & 71.4 & 55.2 \\
Gemini-3-Flash-Preview
                              & \textbf{76.7} & \textbf{70.0} & \textbf{73.8} & 53.2 &  \textbf{77.6} & 40.8 & \textbf{81.2} & 56.0 &  \textbf{78.3} & 55.0 \\
Kimi-K2.5        & 71.7 & 67.2 & 61.4 & \underline{54.4} & \underline{75.3} & \underline{56.4} & 73.0 & 49.2 & 71.4 & 56.8 \\
\midrule
\multicolumn{11}{c}{\textit{API-Based Models (Thinking)}} \\
\midrule
GPT-5.2-Thinking (medium)
                              & \underline{72.3} & \textbf{73.2} & \underline{70.9} & \underline{58.8} & 64.9 & 44.4 & 75.2 & 53.6 & 71.0 & 57.5 \\
DeepSeek-V3.2-Thinking
                              & 70.4 &  69.2 & 60.2 & 49.2 & 75.3 & 54.4 & 77.2 & \textbf{65.6} & 72.8 & 59.6 \\
Gemini-3-Flash-Preview-Thinking
                              & \textbf{75.8} & \underline{70.4} & \textbf{79.7} & \textbf{65.2} & \textbf{81.8} & \textbf{64.0} &  \textbf{83.4} & 63.6 & \textbf{81.6} & \textbf{65.8} \\
Kimi-K2.5-Thinking    & 70.6 & 68.4 & 66.6 & 58.0 & \underline{79.8} & \underline{57.6} & \underline{78.3} & \textbf{65.6} & \underline{75.9} & \underline{62.4} \\

\midrule
\multicolumn{11}{c}{\textit{Open-Source Models (Non-Thinking)}} \\
\midrule
Qwen3-4B              & 58.6 & \underline{59.2} & 35.8 & 29.2 & 65.0 & 29.2 & 57.9 & 38.0 & 55.9 & 38.9 \\
Qwen3-8B          & 60.4 & 57.2 & 39.9 & 32.0 & 64.9 & 30.8 & 58.5 & 42.8 & 57.1 & 40.7 \\
Qwen3-4B-Instruct-2507            & \textbf{66.6} & \underline{59.2} & \underline{44.0} & \textbf{36.0} & \underline{65.8} & \underline{33.6}& 59.2 & \underline{48.8} & \underline{58.9} & \textbf{44.4} \\
Qwen3-30B-A3B-Instruct-2507       & \underline{65.8} & 54.4 & \textbf{48.3} & \underline{35.2} & \textbf{72.4} & \textbf{36.4} & \textbf{67.2} & \textbf{49.6} & \textbf{65.0} & \underline{43.9} \\
LLaMA-3.1-8B-Instruct            & 34.6 & 53.6 & 26.8 & 31.2 & 61.2 & 28.8 & \underline{61.1} & 47.2 & 52.3 & 40.2 \\
LLaMA-3.2-3B-Instruct             & 44.3 & 58.4 & 22.5 & 27.6 & 37.7 & 23.6 & 37.6 & 40.4 & 35.3 & 37.5 \\
LLaMA-3.3-70B-Instruct             & 46.3 & \textbf{60.0} & 36.6 & 30.8 & 64.6 & 30.8 & 56.9 & 46.8 & 54.4 & 42.1 \\

\midrule
\multicolumn{11}{c}{\textit{Open-Source Models (Thinking)}} \\
\midrule
Qwen3-4B-Thinking         & 60.1 & 56.0 & 44.4 & 39.6 & 65.5 & 33.2 & 60.3 & \underline{47.6} & 58.8 & 44.1 \\
Qwen3-8B-Thinking         & 59.7 & 58.0 & \underline{45.9} & \underline{41.2} & \underline{70.9} & \textbf{38.8} & \underline{66.4} & 46.0 & \underline{63.2} & \underline{46.0} \\
Qwen3-4B-Thinking-2507    & \underline{66.9} & \underline{60.4} & 41.9 & 36.8 & 65.4 & 33.6 & 63.0 & 46.8 & 60.0 & 44.4 \\
Qwen3-30B-A3B-Thinking-2507
                                   & \textbf{70.0} & \textbf{64.8} & \textbf{53.1} & \textbf{46.4} & \textbf{73.2} & \underline{35.2} & \textbf{71.8} & \textbf{61.6} & \textbf{68.5} & \textbf{52.0} \\
\bottomrule
\end{tabular*}

\end{table*}

\subsection{Setup}
\paragraph{Evaluated LLMs}
To evaluate step-level process diagnosis, we benchmark 20 models including proprietary API-based models and open-source models. For API-based models, we include GPT-5.2 (Base, Chat, and Thinking), DeepSeek-V3.2 (Non-thinking and Thinking), Gemini-3-Flash-Preview (Minimal and Thinking), and Kimi-K2.5 (Non-Thinking and Thinking). For open-source models, we evaluate the Qwen3 family (4B, 8B, and 30B-A3B) across both standard and thinking variants, as well as the LLaMA-3 series (3.1-8B, 3.2-3B, and 3.3-70B). 
To ensure a fair comparison, we employ a consistent prompt across all experiments (see Appendix~\ref{appendix:prompt}). 
For thinking models, we adopt the recommended sampling parameters, while non-thinking models are evaluated using greedy decoding.

\paragraph{Metrics}
We adopt two complementary metrics to evaluate step-level process quality, targeting global labeling reliability and early error localization.

\textbf{(1) Step Accuracy (StepAcc).}
We compute the micro-averaged agreement ratio between model predictions and human annotations:
\[
\text{StepAcc}
=
\frac{\#\text{matched step labels}}
     {\#\text{all assistant steps}}.
\]
All assistant steps across all trajectories are pooled together, so StepAcc reflects overall step-level labeling quality with longer trajectories contributing proportionally more steps.

\textbf{(2) First-Error Accuracy (FirstErrAcc).}
For each trajectory, we identify the first step labeled as $-1$ and compare its index with human annotations:
\[
\text{FirstErrAcc}
=
\frac{\#\text{samples with matched first-error index}}
     {\#\text{total samples}}.
\]
If neither prediction nor reference contains a $-1$ label, the trajectory is considered error-free and counted as correct.
Unlike StepAcc, FirstErrAcc is less susceptible to error propagation after the first mistake and is not influenced by trajectory length, directly measuring a model’s ability to pinpoint the earliest critical failure~\cite{zheng2025processbench}.
Together, StepAcc captures global process correctness, while FirstErrAcc emphasizes early failure detection in long-horizon trajectories.

\subsection{Main Results}
We present the evaluation results in Table \ref{table:detailedresults}. Our
observations are summarized as follows:

\begin{itemize}[leftmargin=*]
    \item \textbf{Open-source models still lag behind proprietary models.}
    For example, the strongest open-source model, Qwen3-30B-A3B-Thinking-2507, achieves an average {StepAcc} of 68.5\%, whereas the proprietary Gemini-3-Flash-Preview-Thinking attains a substantially higher score of 81.6\%.
    A similar performance gap is observed across individual benchmarks, indicating that the disparity is not limited to a specific task type or evaluation setting.

    \item \textbf{Model scale and reasoning mechanisms are pivotal for accurate step-level evaluation.} 
    As shown in Table \ref{table:detailedresults}, larger model parameters consistently lead to performance gains; for both the Qwen and Llama families, scaling from 3B to 70B improves results across all metrics.   
    However, we can observe that newer models may possess higher capability density~\cite{xiao2025densing}. This is exemplified by Qwen3-4B-Instruct-2507, which, despite having only 50\% of the parameters of the Qwen3-8B model, achieves a superior {FirstErrAcc} (44.4\% vs. 40.7\%).
    In addition, thinking models significantly outperform their instruct counterparts at the same parameter scale. 
    For instance, under a controlled setting with identical parameters, Qwen3-8B in reasoning mode achieves a 6.1\% higher {StepAcc} and a 5.3\% higher {FirstErrAcc} than its non-reasoning variant.    
    Notably, while thinking models generally dominate, GPT-5.2-Chat markedly outperforms its thinking variant on multi-turn tool-use benchmarks (BFCL and $\tau^2$-bench). We hypothesize this is due to specialized optimizations for dialogue dynamics in chat-tuned models, whereas current thinking models may be more tailored for complex single-turn reasoning.

    \item \textbf{Increased task complexity significantly hampers error localization, particularly for smaller models.} 
    The difficulty of identifying critical errors scales with dataset complexity. Moving from HotPotQA to GAIA, almost all models exhibit a performance decline, but the drop is more pronounced for weaker models. For the frontier non-thinking model, Gemini-3-Flash-Preview, {StepAcc} and {FirstErrAcc} decrease by 2.9\% and 16.8\%, respectively. In contrast, Qwen3-4B suffers much sharper declines of 22.8\% and 30\%. This suggests that while large-scale models are more robust, localizing errors in long-horizon, complex tasks remains a significant bottleneck for smaller-scale models.

    \item \textbf{StepAcc and FirstErrAcc are strongly correlated, with first error localization being more challenging.} 
    Across different model families and scales, we observe a strong positive correlation between StepAcc and FirstErrAcc, with an average Pearson $r=0.90$ and Spearman $\rho=0.92$, indicating that models capable of reliable step-level labeling are generally more proficient at identifying the initial critical error.
    However, despite this high correlation, FirstErrAcc is consistently lower than StepAcc, and the gap can be substantial.
    For example, the strongest model Gemini-3-Flash-Preview achieves an average StepAcc of 81.6\% but only 65.8\% FirstErrAcc.
    This systematic discrepancy suggests that accurately localizing the first critical error constitutes a more demanding capability than overall step-level assessment.
    We hypothesize that a single early mistake can induce cascading downstream errors, making the identification of the true root cause substantially harder than recognizing that later steps are incorrect.

\end{itemize}

\subsection{Detailed Analysis}

\paragraph{Dataset-Specific Policy Failure Modes}

\begin{figure}
    \centering
    \includegraphics[width=1\linewidth]{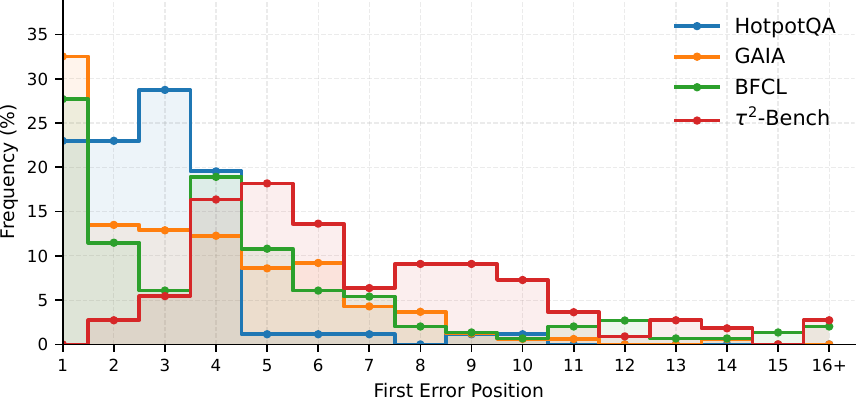}
    \caption{Distribution of first error positions.}

    \label{fig:dist_of_traj_2}
\end{figure}

We observe that different datasets exhibit distinct failure modes, reflecting their heterogeneous task structures and interaction dynamics.
As shown in Figure~\ref{fig:dist_of_traj_2}, the position at which the first error occurs varies substantially across datasets.
For $\tau^2$-Bench, first errors are more likely to appear at later steps in the trajectory.
This suggests that models can advance the task correctly during the initial interactions by invoking appropriate tools, while failures tend to emerge as the interaction progresses, often due to policy violations or difficulties in correctly interpreting newly introduced user requirements.
In contrast, for HotpotQA and GAIA, a significant portion of errors occurs at Step 1. Qualitative analysis suggests this is frequently driven by invalid tool invocations (e.g., syntax or formatting errors) or an immediate failure to formulate a viable information-seeking strategy.

\begin{figure}[t]
    \centering
    \setlength{\tabcolsep}{2pt}

    \begin{tabular}{cc}
        \includegraphics[width=0.5\columnwidth]{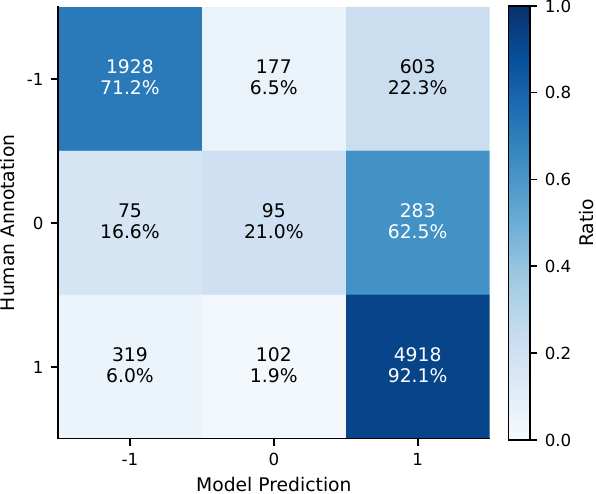} &
        \includegraphics[width=0.5\columnwidth]{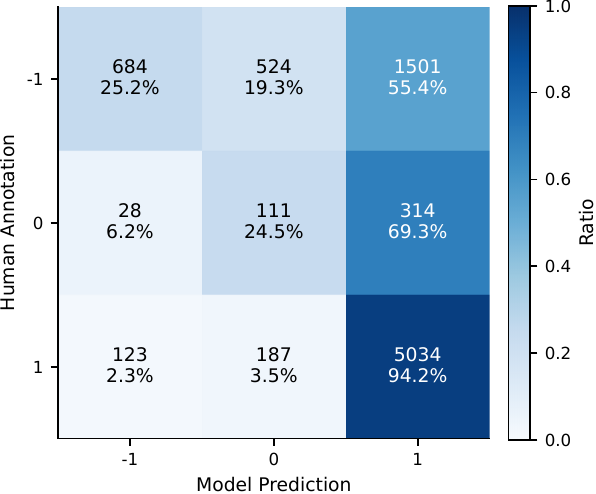} \\[-1pt]

        \raisebox{1pt}{\scriptsize (a) Gemini-3-Flash-Preview} &
        \raisebox{1pt}{\scriptsize (b) Qwen3-30B-A3B-Thinking}
    \end{tabular}

    \caption{
    Row-normalized confusion matrices of step evaluations for Gemini-3-Flash-Preview and Qwen3-30B-A3B-Thinking.
    }
    \label{fig:confusion-matrix}
\end{figure}

\paragraph{PRMs Struggle to Distinguish Neutral and Incorrect Steps.}
We visualize the step-level confusion matrices in Figure~\ref{fig:confusion-matrix}.
\textbf{First}, both the strongest closed-source and open-source models exhibit a tendency to over-predict the positive ($+1$) label, as evidenced by substantial probability mass on the $+1$ prediction column across all rows.
This bias is more pronounced for Qwen3-30B-A3B-Thinking, leading to a higher rate of false positives, which is the main cause for its lower overall accuracy compared to Gemini-3-Flash-Preview.
\textbf{Second}, neutral ($0$) steps are consistently harder than positive and negative steps for both models: the confusion mass for the $0$ row is more dispersed, and misclassifications frequently collapse to the positive label.
We attribute this primarily to the inherent ambiguity of neutral labels.
Unlike clearly correct or clearly harmful actions, the utility of many neutral steps is context-dependent and often only becomes evident through downstream effects (e.g., whether the retrieved evidence is later used, whether uncertainty is reduced, or whether alternative actions would have sufficed).
Consequently, neutral steps exhibit a weak and delayed supervisory signal, making them inherently harder to judge from the local step alone and thus more prone to misclassification.
More qualitative error analysis of PRMs can be found in Appendix~\ref{sec:qualitative}.

\begin{figure}
    \centering
    \includegraphics[width=1\linewidth]{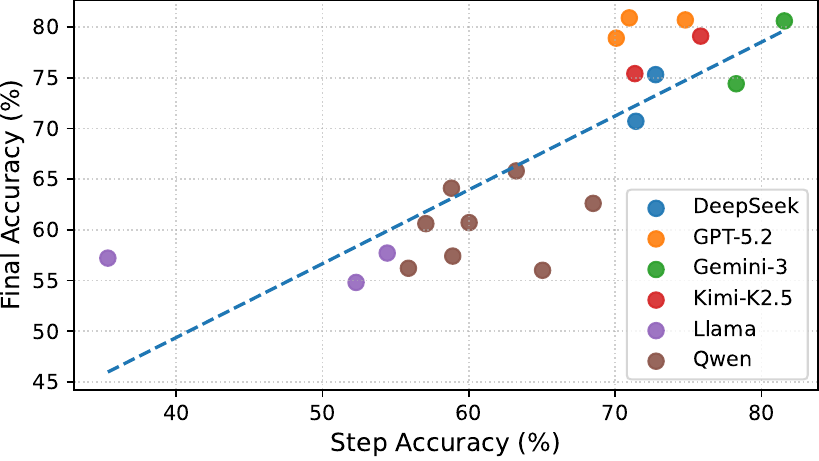}
    \caption{Correlation between step-level evaluation accuracy and trajectory-level final accuracy. 
    It shows high correlation (Pearson $r=0.814, p=1.2\times10^{-5}$).  
    }

    \label{fig:step-vs-final}
\end{figure}

\paragraph{Stronger ORMs tend to be stronger PRMs.}
Figure~\ref{fig:step-vs-final} shows a clear positive association between step-level evaluation accuracy (\textit{StepAcc}) and trajectory-level final accuracy across all evaluated models. The correlation is strong and statistically significant (Pearson $r=0.814$, $p=1.2\times 10^{-5}$), indicating that models that are better at predicting step correctness also tend to be more reliable at predicting final outcomes. 
Consistent with this trend, top-performing models such as GPT-5.2 and Gemini-3 concentrate in the upper-right region, while smaller models largely fall in the lower-left.
Despite this strong global correlation, StepAcc is not a redundant proxy for outcome performance: some models with similar final accuracy exhibit markedly different step accuracy (e.g., Llama vs.\ Qwen around $\sim$57\% final accuracy), suggesting residual variation beyond a purely outcome-based view. 
This residual variation highlights the unique value of process supervision and motivates future research on developing highly capable yet parameter-efficient process reward models.

\paragraph{Process-derived signals improve Best-of-$N$ selection.}
Table~\ref{tab:bon-selection} compares outcome-based and process-based heuristics for Best-of-$N$ sampling.
We observe that simple step-level positivity statistics (\#\,Pos and \%\,Pos) are effective test-time scaling strategies and often outperform the outcome-only criterion for weaker generators such as Qwen3-30B-A3B.
Furthermore, combining outcome- and process-level signals in a two-stage selector consistently improves performance across all generators.
This suggests that process-derived scores provide complementary discriminative signals for refining or tie-breaking among ORM-selected candidates.
Nevertheless, the oracle Pass@$N$ upper bound ($77.4\%$) remains substantially higher than all practical Best-of-$N$ strategies evaluated here, indicating considerable headroom for future reward models.


\begin{table}[t]
\small
\centering
\setlength{\tabcolsep}{3pt}
\renewcommand{\arraystretch}{1.05}
\caption{
Comparison of Best-of-$N$ ($N{=}8$) selection strategies on GAIA.
\textit{Final} selects the first trajectory predicted to be successful at the outcome level.
\#\,Pos Step and \%\,Pos Ratio select trajectories based on the number and proportion of positive steps, respectively.
The two-stage strategy (\textit{Final$\rightarrow$Process}) combines outcome- and process-level criteria by refining outcome-based filtering with step-level signals.
As references, majority-voting achieves 49.1\% accuracy, while the oracle Pass@$N$ is 77.4\%.
}

\label{tab:bon-selection}
\begin{tabular}{lcccc}
\toprule
\textbf{Generator} &
\textbf{Final} &
\textbf{\#\,Pos } &
\textbf{\%\,Pos } &
\textbf{Two-Stage} \\
\midrule
Qwen3-30B-A3B-Instruct-2507  & 37.7 & 41.5 & 47.2 & 43.4 \\
Qwen3-30B-A3B-Thinking-2507  & 35.9 & 45.3 & 49.1 & 50.9 \\
DeepSeek-V3.2 & 47.2 & 37.7 & 45.3 & 49.1 \\
DeepSeek-V3.2-Thinking       & \textbf{56.6} & \textbf{50.9} & \textbf{54.7} & \textbf{64.2} \\
Gemini-3-Flash-Preview & \textbf{56.6} & \underline{49.1} & \textbf{54.7} & \underline{58.5} \\
\bottomrule
\end{tabular}
\end{table}

\begin{table}[!t]
\centering
\caption{
Ablation of error propagation in step-level evaluation protocols for GAIA Best-of-$N$ ($N{=}8$) selection.
All scores report using the two-stage (\textit{Final$\rightarrow$Process}) selector.
\textit{Original} uses the error-propagation annotation protocol, whereas \textit{w/o EP} evaluates each step independently without propagating errors.
$\Delta$ denotes the change from \textit{Original} to \textit{w/o EP}.
}
\label{tab:error_propagation_ablation}
\small
\setlength{\tabcolsep}{6pt}
\begin{tabular}{lccc}
\toprule
\textbf{Step Evaluator} & \textbf{Original} & \textbf{w/o EP} & $\Delta$ \\
\midrule
DeepSeek-V3.2 & 49.1 & 52.8 & $+3.7$ \\
DeepSeek-V3.2-Thinking & 64.2 & 50.9 & $-13.2$ \\
Gemini-3-Flash-Preview & 58.5 & 52.8 & $-5.7$ \\
\bottomrule
\end{tabular}
\end{table}

\paragraph{Ablation on Error-Propagation Rules}
We compare the original error-propagation annotation protocol with an independent-judgment variant, in which each step is evaluated independently without propagating errors from preceding steps.
As shown in Table~\ref{tab:error_propagation_ablation}, removing error propagation improves the two-stage Best-of-8 accuracy when using DeepSeek-V3.2 as the step evaluator, increasing the score from 49.1 to 52.8. However, it substantially degrades performance for stronger models, reducing DeepSeek-V3.2-Thinking from 64.2 to 50.9 and Gemini-3-Flash-Preview from 58.5 to 52.8. These results suggest that error propagation can benefit test-time scaling when the step evaluator is capable of reliably identifying error cascades.

\begin{figure}[t]
    \centering
    \includegraphics[width=\linewidth]{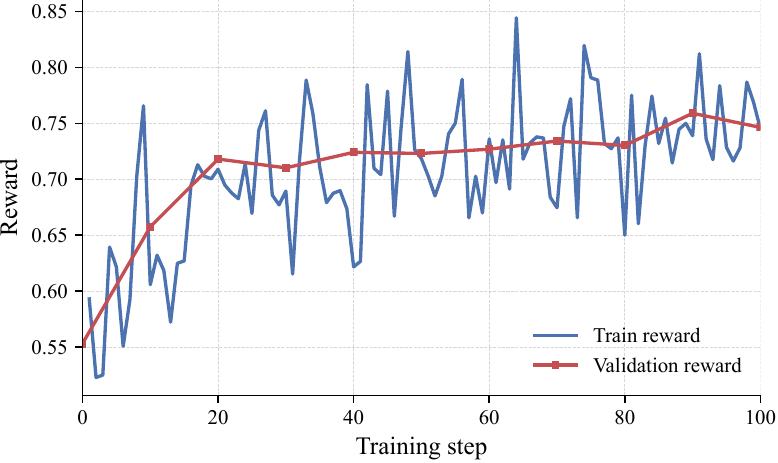}
    \caption{StepAcc on 800 training samples and 200 validation samples over the course of RL training.}
    \label{fig:rl}
\end{figure}

\paragraph{Training on \bench}
Although \bench is primarily designed for evaluation, its step-level annotations also enable lightweight reinforcement learning experiments. 
To examine this auxiliary use case, we conduct a GRPO training experiment on 800 annotated trajectories, using StepAcc as the reward signal, and evaluate the resulting model on the other 200 trajectories.
As shown in Figure~\ref{fig:rl}, StepAcc increases consistently on both the training and validation sets over the course of RL training. In particular, validation StepAcc improves from 55.3\% to 74.6\%, suggesting that \bench provides useful process-level supervision beyond serving as a diagnostic evaluation benchmark.

\paragraph{Effect of Anchoring Bias}
During annotation, we provided annotators with predictions from three state-of-the-art models as references, which may introduce anchoring bias. To assess this effect, we re-annotated 100 randomly sampled trajectories without model-generated references using a separate group of annotators. The step-level agreement with the original labels is 84.06\%, compared with the inter-annotator agreement of 89.1\% in the main study. This result suggests that the references are more likely to reduce annotation difficulty than to introduce substantial systematic bias.


\section{Conclusion}

In this paper, we introduce \bench, the first human-annotated benchmark designed to evaluate the effectiveness of intermediate steps in tool-using agents.
Our extensive evaluation across 20 LLMs yields several pivotal insights. 
First, we observe that closed-source models and thinking models consistently outperform their counterparts.
Second, we find that while models are increasingly capable, they may exhibit a notable bias toward positive labels and struggle to distinguish neutral steps. 
Third, our experiments demonstrate that process-derived signals provide complementary value to outcome supervision, achieving better Best-of-$N$ performance.
We envision \bench as a cornerstone testbed for tool-using process reward models, catalyzing future research towards more powerful and reliable agentic systems.
As future work, we plan to extend \bench to additional domains, such as GUI-based agents and computer-using agents.

\section*{Acknowledgements}
This work was supported by the National Key R\&D Program of China (No.\ 2024YFC3306500), the National Natural Science Foundation of China (No.\ 62376273) and the Beijing Nova Program (No. 20240484568).

\bibliographystyle{ACM-Reference-Format}
\bibliography{ref}  

\appendix

\section{Limitations}
Despite careful efforts (Section~\ref{sec:benchmark_construction}), \bench may still be affected by annotation noise and inherent human subjectivity.
In addition, the current version of \bench is restricted to text-only tool use and does not yet capture multimodal agent behaviors, such as GUI-based interaction or
navigation. Extending \bench to these more complex
multimodal settings is an important direction for future work.




\section{PRM Evaluation Prompt} \label{appendix:prompt}

In this section, we provide the comprehensive prompt template used for evaluating models as \textbf{Process Reward Models (PRMs)} in Figure \ref{fig:PRM_Evaluation_Prompt_Template}. The prompt instructs the model to act as \textbf{a strict but fair trajectory annotator for tool-use agents}. It defines a granular three-tier scoring schema: \textbf{$+1$ (Correct and effective)} for steps that move the task forward; \textbf{$0$ (Neutral or exploratory)} for reasonable but non-pivotal steps, such as redundant reasoning or failed tool calls due to external factors; and \textbf{$-1$ (Incorrect or harmful)} for steps containing factual errors, tool misuse, or constraint violations. Furthermore, the template incorporates a set of rigorous logical constraints, such as the "cumulative penalty" rule for subsequent steps following an error and the "hindsight bias avoidance" principle. This structured prompt ensures that the evaluation process is both consistent and logically sound across diverse task scenarios.

\begin{figure*}[t]
\centering
    \begin{blueexampleFloatingTitle}{PRM Evaluation Prompt Template}

        \promptsection{System Message}            
        
        You are a strict but fair trajectory annotator for tool-use agents.

        You will be given one complete trajectory consisting of system, user, assistant, and tool messages, together with the tool definitions.

        Your task is to label EACH assistant message (each assistant message constitutes one Step) using the following scheme:

        \begin{itemize}
            \item \textbf{$+1$: Correct and effective.} The step is factually correct given the information available at that time and clearly moves the task closer to successful completion by: (i) correctly invoking a tool or interpreting tool outputs, or (ii) introducing valid constraints, decisions, or information that reduces the remaining uncertainty of the task.
            \item \textbf{0: Neutral or exploratory.} The step is reasonable but has limited or unclear impact on task progress. This includes exploratory reasoning, redundant restatements, partial planning, or cases where the correctness is debatable given the available evidence. Tool calls that fail due to external reasons (e.g., timeout, 404), when the attempt itself is reasonable, are typically labeled 0.
            \item \textbf{$-1$: Incorrect or harmful.} The step contains factual errors, misinterprets tool outputs, violates constraints, repeats failed actions without a meaningful change in strategy, fabricates tool results or evidence, or otherwise pushes the trajectory away from successful completion.
        \end{itemize}

        \textbf{Important rules:}

        \begin{itemize}
            \item Only assistant messages are labeled. User and tool messages serve only as evidence.
            \item Avoid hindsight bias: judge each step strictly based on the information available up to that point in the trajectory.
            \item Any step labeled $-1$ triggers a cumulative penalty: all subsequent assistant steps in the same workflow should also be labeled $-1$, unless one of the following holds: (i) the assistant explicitly acknowledges and corrects the earlier mistake, or (ii) the assistant produces a subsequent step that no longer depends on the incorrect assumption and effectively resumes progress toward the task.
            \item Repeating the same failed action without a meaningful change in parameters or strategy typically transitions from 0 to $-1$.
            \item If an incorrect statement does not affect any subsequent reasoning or actions and is not relied upon later, it may be labeled 0; otherwise, it should be labeled $-1$.
            \item Any violation of the policies or requirements specified in the system prompt results in a score of $-1$, except for certain output-formatting norms. The following behaviors are considered acceptable and do not incur penalties: providing a text response simultaneously with a tool call, not conducting reasoning before a tool call, failing to encapsulate reasoning content within \texttt{<think>...</think>} tags, responding to the user while executing a function call, or executing multiple parallel tool calls.
            \item A score of $+1$ is assigned if the entire conversation is initiated by the assistant and its first message is a greeting; this exemption applies only to the first message.
            \item Upon user request, if the assistant executes specific instructions, a score of $+1$ shall be awarded, notwithstanding any deviation from the overarching objective.
        \end{itemize}

        After labeling all assistant steps, also assign a label to:

        \textbf{FINAL\_RESULT:} \\
        $+1$: The overall task is successfully completed. \\
        $-1$: The task fails due to incorrect reasoning, tool misuse, or unresolved errors.

        \textbf{Output format:}

        You MUST first provide your reasoning process, analyzing each assistant step one by one. Then, at the very end, output a JSON object wrapped in \texttt{```json ... ```} markdown code block as your judgement results.

    \promptsection{User Message}

{\small\ttfamily \noindent\{\\ \hspace*{1em}"question": QUESTION,\\ \hspace*{1em}"task\_description": TASK\_DESCRIPTION,\\ \hspace*{1em}"tools": TOOL\_DEFINITIONS,\\ \hspace*{1em}"messages": [\\ \hspace*{2em}[MESSAGE\_INDEX, MESSAGE\_OBJECT],\\ \hspace*{2em}...\\ \hspace*{1em}],\\ \hspace*{1em}"assistant\_message\_indices": [ASSISTANT\_MESSAGE\_INDEX, ...],\\ \hspace*{1em}"notes": \{\\ \hspace*{2em}"step\_definition": "Each Step corresponds to one message whose role is assistant. Use the provided message indices.",\\ \hspace*{2em}"output\_requirements": "Return JSON with step\_labels, final\_label, and explanations."\\ \hspace*{1em}\}\\ \noindent\} }
    \vspace{-0.3em}
        
    \end{blueexampleFloatingTitle}
    \vspace{-1em} 
    \captionof{figure}{PRM Evaluation Prompt Template} 
    \label{fig:PRM_Evaluation_Prompt_Template}
\end{figure*}

\section{Analysis on Long-Horizon Trajectories}
We analyze long-horizon cases by evaluating the top 25\% longest trajectories in each dataset. This analysis leads to three observations. First, all models show clear performance degradation, with StepAcc decreasing by 3.8--13.49 points and FirstErrAcc decreasing by 10.79--21.67 points, confirming the increased difficulty of longer trajectories. Second, FirstErrAcc drops more substantially than StepAcc across models, suggesting that first-error localization is particularly sensitive to trajectory length. Third, frontier models are generally more robust than smaller open-source models. An interesting exception is Gemini: the non-thinking variant suffers substantial degradation, whereas the thinking-enabled variant remains relatively stable, suggesting that explicit reasoning may help preserve robustness on long-horizon trajectories.

\section{Qualitative Error Analysis} \label{sec:qualitative}

To further elucidate the limitations of current LLMs in trajectory evaluation, we categorize the most prevalent errors into the following five dimensions:

\begin{itemize}[topsep=1pt, partopsep=1pt, itemsep=2pt, leftmargin=15pt]
\item \textbf{Negligence of Informational Errors} refers to the failure of LLMs to precisely detect factual inaccuracies or violations of predefined policy constraints within trajectory messages, especially when these errors are small or hidden in long texts.

\item \textbf{Negligence of Logical Reasoning Errors} refers to the model's inability to identify flaws in the underlying reasoning chain, such as the repetition of failed actions without strategic adjustments or the presence of causal inconsistencies in the agent's decision-making process.

\item \textbf{Negligence of Tool Invocation Errors} refers to cases where LLMs struggle to pinpoint nuanced issues in tool calling, including improper tool selection, syntax malformations, or indirect misuse, which tests the model's power to distinguish valid invocations from invalid ones.

\item \textbf{Misjudgment of Correct Steps via Overthinking} refers to the tendency of LLMs to mislabel otherwise correct steps by over-interpreting rules or imposing imaginary constraints on the context, often resulting in excessive stringency and false negatives in evaluation.

\item \textbf{Boundary Ambiguity for Neutral or Exploratory Steps} refers to the LLM's difficulty in identifying the appropriate scope of exploratory behaviors, failing to distinguish among acceptable exploration ($0$), redundant attempts leading to inefficiency ($-1$), and essential exhaustive search processes ($+1$).
\end{itemize}

\section{Ethical Statement}

Throughout the entire process of constructing AgentProcessBench, we strictly adhere to ethical standards concerning data privacy, human labor, and the broader impact of autonomous agents. The benchmark is constructed using tasks from four widely recognized public datasets: HotpotQA, GAIA, BFCL, and $\tau^2$-bench. All interaction trajectories were generated by Large Language Models (LLMs), ensuring that the dataset contains no private or sensitive personal information (PII) from real human-to-human interactions. By utilizing these rollouts, we provide a realistic evaluation environment while completely mitigating privacy risks associated with user data.

Regarding human annotation, we recruited experts with at least undergraduate-level training in computer science and a minimum of one year of experience with LLMs. All annotators underwent a mandatory proficiency test and a specialized tutorial to ensure they understood the ternary labeling scheme and error-propagation rules. To uphold fair labor practices, all participants were compensated at a competitive rate exceeding local standards for professional technical work. To guarantee the reliability and objectivity of the labels, each trajectory was independently reviewed by two experts, with an inter-annotator agreement of 89.1\%. Any remaining discrepancies were resolved through collective discussion to reach a final consensus.

The primary motivation for this research is to advance the development and evaluation of Process Reward Models (PRMs) tailored for tool-augmented agents. Unlike outcome-based metrics, AgentProcessBench provides a rigorous, human-verified testbed to evaluate how accurately models can diagnose the effectiveness of individual steps within complex, open-world trajectories. This capability is critical because tool-use failures frequently involve irreversible side effects—such as deleting essential files—making the precise evaluation of step-level verification models a safety priority. By establishing standardized criteria for identifying harmful actions (-1) and providing dense step-level annotations, this benchmark aims to foster the creation of more reliable reward models that can eventually guide agents toward safer and more intent-aligned behaviors.

During the preparation of this work, the authors used LLMs to improve the language and grammar of the manuscript. After using this tool, the authors reviewed and edited the content as needed. The authors take full responsibility for the accuracy and integrity of the paper's content.

\end{document}